\documentclass{llncs}
\usepackage{graphicx}
\usepackage[utf8]{inputenc}
\usepackage{subcaption}
\usepackage{fontawesome}
\usepackage{booktabs}
\usepackage{soul}  
\usepackage{graphicx}
\usepackage[utf8]{inputenc}
\usepackage{color}
\usepackage{latexsym}
\usepackage{comment}
\usepackage{enumitem}
\usepackage{hyperref}
\usepackage[colorinlistoftodos]{todonotes}
\usepackage{multirow}
\usepackage{array}
\newcolumntype{?}{!{\vrule width 1pt}}

\newcommand{\red}[1]{{\color{red} #1}}

\newcommand{\ct}{\texttt{CheckThat!}}

\begin{document}
\title{Overview of the CLEF-2019 CheckThat! Lab: Automatic Identification and Verification\\ of Claims}

%
\titlerunning{Overview of the CLEF-2019 CheckThat! Lab}
%
\author{%
Tamer Elsayed\inst{1}
\and 
Preslav Nakov\inst{2}
\and  
Alberto Barr\'{o}n-Cede\~no\inst{3}
\and
Maram Hasanain\inst{1}
\and \\	 
Reem Suwaileh\inst{1}
\and 
Giovanni Da San Martino\inst{2}
\and
Pepa Atanasova\inst{4}
}

\authorrunning{T. Elsayed et al.}

\institute{Computer Science and Engineering Department, Qatar University, Doha, Qatar \\
\email{\{telsayed, maram.hasanain, rs081123\}@qu.edu.qa}\and
Qatar Computing Research Institute, HBKU, Doha, Qatar \\
\email{\{pnakov, gmartino\}@qf.org.qa}\\\and
DIT, Università di Bologna, Forlì, Italy \\
\email{a.barron@unibo.it}\\\and
Department of Computer Science, University of Copenhagen, Denmark \\
\email{pepa@di.ku.dk}\\
}

\maketitle              

\setcounter{footnote}{0}
\begin{abstract}
We present an overview of the second edition of the \ct\ Lab at CLEF 2019. The lab featured two tasks in two different languages: English and Arabic. Task~1 (English) challenged the participating systems to predict which claims in a political debate or speech should be prioritized for fact-checking. Task~2 (Arabic) asked to (A)~rank a given set of Web pages with respect to a check-worthy claim based on their usefulness for fact-checking that claim, (B)~classify these same Web pages according to their degree of usefulness for fact-checking the target claim, (C)~identify useful passages from these pages, and (D) use the useful pages to predict the claim's factuality. \ct \ provided a full evaluation framework, consisting of data in English (derived from fact-checking sources) and Arabic (gathered and annotated from scratch) and evaluation based on mean average precision (MAP) and normalized discounted cumulative gain (nDCG) for ranking, and F$_1$ for classification. 
A total of 47 teams registered to participate in this lab, and fourteen of them actually submitted runs (compared to nine last year). The evaluation results show that the most successful approaches to Task~1 used various neural networks and logistic regression. As for Task~2, learning-to-rank was used by the highest scoring runs for subtask A, while different classifiers were used in the other subtasks. We release to the research community all datasets from the lab as well as the evaluation scripts, which should enable further research in the important tasks of check-worthiness estimation and automatic claim verification.

\end{abstract}


\section{Introduction}

With the rise of ``fake news,'' which spread in all types of online media, the need arose for systems that could detect them automatically~\cite{rubin2015deception}. The problem has various aspects~\cite{shu2017fake}, but here we are interested in predicting which claims are worth fact-checking, what information is useful for fact-checking, and finally predicting the factuality of a given claim~\cite{baly2018integrating,NAACL2018:claimrank,Nie19,popat2018declare,thorne2018fever,yoneda2018ucl}. 
Evidence-based fake news detection systems can serve fact-checking in two ways: 
(\emph{i})~by facilitating the job of a human fact-checker, but not replacing her, and 
(\emph{ii})~by increasing her trust in a system’s decision~\cite{nguyen2018interpretable,popat2018declare,thorne2018fever}. We focus on the problem of checking the factuality of a claim, which has been studied before but rarely in the context of evidence-based fake news detection systems~\cite{Atanasova:2019:AFU:3331015.3297722,ba2016vera,castillo2011information,Hardalov2016,RANLP2017:clickbait,RANLP2017:factchecking:external,ma2016detecting,mihaylova-etal-2019-semeval,AAAI2018:factchecking,mukherjee2015leveraging,popat2016credibility,Yasser2018:CIKM,zubiaga2016analysing}. 

There are several challenges that make the development of automatic fake news detection systems difficult:
\begin{enumerate}
 \item \label{it:huge}
 A fact-checking system is effective if it is able to identify a false claim before it reaches a large number of people. Thus, the current speed at which claims spread on the Internet and social media imposes strict efficiency constraints on fact-checking systems.
 \item \label{it:diff} The problem is difficult to the extent that, in some cases, even humans can hardly distinguish between fake and true news~\cite{shu2017fake}.
 \item \label{it:bench}
 There are very few \emph{large-scale} benchmark datasets that could be used to test and improve fake news detection systems~\cite{shu2017fake,thorne2018fever}.
 \end{enumerate}
 
\begin{figure}[h]
\centering
\includegraphics[width=1.0\columnwidth]{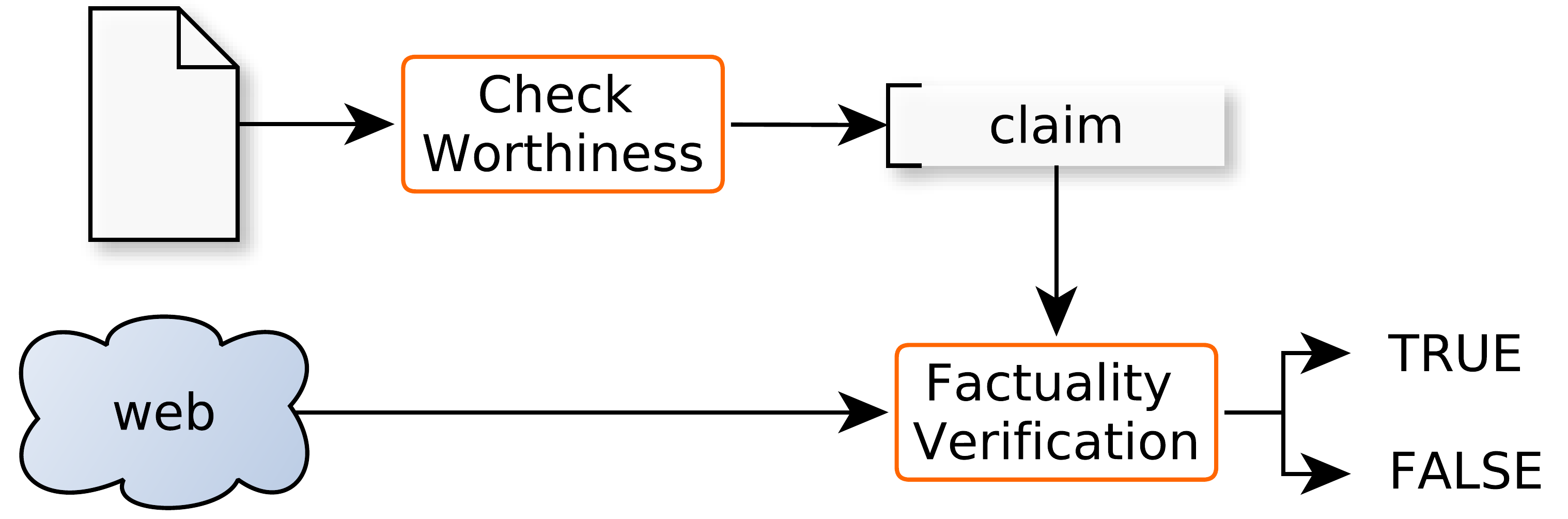}
\vspace*{-1mm}
\caption{Information verification pipeline with the two tasks in the \ct \ lab: check-worthiness estimation and factuality verification.}
\label{fig:pipeline}
\end{figure}

Thus, in 2018 we started the \ct \ lab on Automatic Identification and Verification of Political Claims \cite{clef2018checkthat:task1,clef2018checkthat:task2,clef2018checkthat:overall}.
Given the success of the lab, we organized a second edition of the lab in 2019 \cite{CheckThat:ECIR2019}, which aims at providing a full evaluation framework along with large-scale evaluation datasets. The lab this year is organized around two different tasks, which correspond to the main blocks in the verification pipeline, as depicted in Figure~\ref{fig:pipeline}.

\noindent
\textbf{Task~1: Check-worthiness Estimation}. The task asks the participating systems to propose which claims in a text should be prioritized for fact-checking. Whereas we focus on transcribed political debates and speeches, the task can be applied to other texts, such as news articles, blog entries, or interview transcriptions. 
Task~1 addresses a problem that is not as well studied as other steps in the fact-checking pipeline~\cite{clef2018checkthat:task1,gencheva-EtAl:2017:RANLP,Hassan:15}. 

\vspace{12pt}
\noindent
\textbf{Task~2: Evidence and Factuality}. This task focuses on extracting evidence from the Web to support the making of a veracity judgment for a given target claim. 
We divide Task~2 into the following four subtasks: (A)~ranking Web pages with respect to a check-worthy claim based on their potential usefulness for fact-checking that claim; 
(B)~classifying Web pages according to their degree of usefulness for fact-checking the target claim; (C)~extracting passages from these Web pages that would be useful for fact-checking the target claim; and (D) using these useful pages to verify whether the target claim is factually true or not. 
\vspace{12pt}

Whereas a practical system should be able to address all these tasks, we propose each of them independently in order to ease participation, to have focused evaluation for each task, and to enable more meaningful comparisons between the participating systems.

The dataset for Task 1 is an extension of the CT-CWC-18 dataset~\cite{clef2018checkthat:task1}, which we used in the 2018 edition of the \ct \ lab. We added annotations from three press conferences, six public speeches, six debates, and one post, all fact-checked by human experts on \url{factcheck.org}. We further manually refined the annotations to make sure they contain only sentences that were part of the fact-checked claim, and to include all occurrences of a claim. For Task~2, we built a new dataset from scratch by manually curating claims, retrieving Web pages through a commercial search engine, and then hiring both in-house and crowd annotators to collect judgments for the four subtasks. As a result of our efforts, we release two datasets. CT19-T1 includes English claims and judgments for Task 1. CT19-T2 includes Arabic claims and retrieved Web pages along with three sets of annotations for the four subtasks. 

A total of fourteen teams participated in this year's lab, which represents about 55\% increase in participation with respect to the 2018 edition~\cite{clef2018checkthat:overall}. In both tasks, the most successful systems relied on supervised machine learning models for both ranking and classification. We believe that there is still large room for improvement, and thus we release the annotated corpora and the evaluation scripts, which should enable further research on check-worthiness estimation and automatic claim verification.\footnote{\url{http://sites.google.com/view/clef2019-checkthat/datasets-tools}}

The remainder of this paper is organized as follows. 
Sections~\ref{sec:task1} and~\ref{sec:task2} discuss Task 1 and Task 2, respectively, covering the task definitions, the evaluation framework, an overview of the participants' approaches, and the official results. Then, Section~\ref{sec:conclusions} draws some conclusions and points to possible directions for future work.

\section{Overview of Task 1: Check-Worthiness}
\label{sec:task1}

\noindent
Task 1 aims at helping fact-checkers prioritize their efforts. In particular, it asks participants to build systems that can mimic the selection strategies of a particular fact-checking organization: \url{factcheck.org}. 
The task  is defined as follows: 
\begin{center}
\parbox[c]{0.85\textwidth}{%
\textit{Given a political debate, interview, or speech, transcribed  and segmented into sentences, rank the sentences with respect to the priority with which they should be fact-checked.}
}
\end{center}

This is a ranking task and the participating systems are asked to produce one score per sentence, according to which the sentences are to be ranked. 
This year, Task~1 was offered for English only (it was also offered in Arabic in 2018 \cite{clef2018checkthat:task1}). 
Figure~\ref{ex:task1} shows examples of annotated debate fragments. In Figure~\ref{ex:task1-a}, Hillary Clinton discusses the performance of her husband Bill Clinton as US president. Donald Trump fires back with a claim that is worth fact-checking, namely that Bill Clinton approved NAFTA\@. In Figure~\ref{ex:task1-b}, Donald Trump is accused of having filed for bankruptcy six times, which is also a claim that is worth fact-checking.
In Figure~\ref{ex:task1-c}, Donald Trump claims that border walls work.
In a real-world scenario, the intervention by Donald Trump in Figure~\ref{ex:task1-a}, the second one by Hillary Clinton in Figure~\ref{ex:task1-b}, and the first two by Donald Trump in Figure~\ref{ex:task1-c} should be ranked on top of the ranked list in order to get the attention of the fact-checker. 

\begin{figure}[t]
\centering
\begin{subfigure}{\textwidth}
\centering
\begin{tabular}{p{17mm} p{88mm} @{\hspace{5mm}}p{5mm}}
\toprule
H.\ Clinton:	& I think my husband did a pretty good job in the 1990s.	& 	\\
H.\ Clinton:	& I think a lot about what worked and how we can make it work again\ldots	& 	\\
D.\ Trump:	& Well, he approved NAFTA\ldots	& \faCheckCircleO	\\
\bottomrule
\end{tabular}         
\caption{Fragment from the First 2016 US Presidential Debate.\label{ex:task1-a}}
\end{subfigure}
\vspace{12pt}

\begin{subfigure}{\textwidth}
\centering
\begin{tabular}{p{17mm} p{88mm} @{\hspace{5mm}} p{5mm}}
\toprule
H.\ Clinton:	& He provided a good middle-class life for us, but the people he worked for, he expected the bargain to be kept on both sides. 	 &	\\
H.\ Clinton:	& And when we talk about your business, you've taken business bankruptcy six times. &	\faCheckCircleO	\\\bottomrule
\end{tabular}         
\caption{Another fragment from the First 2016 US Presidential Debate. \label{ex:task1-b}}
\end{subfigure}
\vspace{12pt}

\begin{subfigure}{\textwidth}
\centering
\begin{tabular}{p{17mm} p{88mm} @{\hspace{5mm}} p{5mm}}
\toprule
D.\ Trump:	& It’s a lot of murders, but it’s not close to 2,000 murders right on the other side of the wall, in Mexico. 	 &	\faCheckCircleO	\\
D.\ Trump:	& So everyone knows that walls work. &	\faCheckCircleO	\\
D.\ Trump:	& And there are better examples than El Paso, frankly. &	\\\bottomrule
\end{tabular}         
\caption{Fragment from Trump’s National Emergency Remarks in February 2019. \label{ex:task1-c}}
\end{subfigure}

 \caption{\label{ex:task1}English debate fragments: check-worthy sentences are marked with \faCheckCircleO.}
 \end{figure}

\subsection{Dataset}
The dataset for Task 1 is an extension of the CT-CWC-18 dataset~\cite{clef2018checkthat:task1}. The full English part of CT-CWC-18 (training and test) has become the training data this year. For the new test set, we have produced labeled data from three press conferences, six public speeches, six debates, and one post. 

As in last year, the annotations for the new instances have been derived from the publicly available analysis carried out by \url{factcheck.org}. We considered those claims whose factuality was challenged by
the fact-checkers as check-worthy and we made them positive instances in CT19-T1. Note that our annotation is at the sentence level. Therefore, if only part of a sentence was fact-checked, we annotated the entire sentence as a positive instance. If a claim spanned more than one sentence, we annotated all these sentences as positive. Moreover, in some cases, the same claim was made multiple
times in a debate/speech, and thus we annotated all these sentences that referred to it rather than only the one that was fact-checked. Finally,
we manually refined the annotations by moving them to a neighboring sentence
(e.g., in case of argument) or by adding/excluding some annotations.
Table~\ref{tab:task1datasets} shows some statistics about the CT19-T1 corpus.

\begin{table}[t]
\centering
\caption{\label{tab:datasets-overview} Total number of sentences and number of check-worthy ones in the CT19-T1 corpus.}
\begin{tabular}{l@{\hspace{9mm}} l@{\hspace{6mm}} r@{\hspace{6mm}}r@{\hspace{6mm}}r @{\hspace{6mm}}r}
\toprule 
 \bf Type & \bf Partition & \bf Sentences& \bf Check-worthy \\ 
 \midrule 
 \multirow{2}{*}{\bf Debates}& Train & 10,648 &  256\\  
 &Test &  4,584 &  46\\  
\midrule
\multirow{2}{*}{\bf Speeches}& Train & 2,718 & 282\\  
 &Test &  1,883 & 50\\  
\midrule
\multirow{2}{*}{\bf Press Conferences}& Train & 3,011 & 36\\  
 &Test & 612 &  14\\  
\midrule
\bf Posts & Train & 44 & 8\\  
\midrule
\multirow{2}{*}{\bf Total}&  Train & \bf 16,421 & \bf 433\\  
 & Test & \bf 7,079 &  \bf 110\\ 

\bottomrule
\end{tabular}
\label{tab:task1datasets}
\end{table}

\subsection{Overview of the Approaches}

Eleven teams took part in Task~1. The most successful approaches relied on training supervised classification models to assign a check-worthiness score to each of the sentences. Some participants tried to model the context of each sentence, e.g.,~by considering the neighbouring sentences to represent an instance~\cite{checkthat19:Favano,hansen2019neural}. Yet, the most successful systems considered each sentence in isolation. Table~\ref{tab:overviewt1} shows an overview of the approaches. Whereas many approaches opted for embedding representations, feature engineering was also popular in this year's submissions. 

The most popular features were bag-of-words representations, part-of-speech (PoS) tags, named entities (NEs), sentiment analysis, and statistics about word use, to mention just a few. Two of the systems also made use of co-reference resolution. The most popular classifiers included SVM, linear regression, Na\"ive Bayes, decision trees, and neural networks.

Team \textbf{Copenhagen} achieved the best overall performance by building upon their approach from 2018~\cite{hansen2019neural,checkthat19:Hansen}. Their system learned dual token embeddings ---domain-specific word embeddings and syntactic dependencies---, and used them in an LSTM recurrent neural network. They further pre-trained this network with previous Trump and Clinton debates, and then supervised it weakly with the ClaimBuster system.\footnote{\url{http://idir.uta.edu/claimbuster/}}
In their primary submission, they further used a contrastive ranking loss, which was excluded from their contrastive1 submission. For their contrastive2 submission, they concatenated the sentence representations for the current and for the previous sentence. 

Team \textbf{TheEarthIsFlat}~\cite{checkthat19:Favano} trained a feed-forward neural network with two hidden layers, which takes as input Standard Universal Sentence Encoder (SUSE) embeddings~\cite{cer2018universal} for the current sentence as well as for the two previous sentences as a context. In their contrastive1 run, they replaced the embeddings with the Large Universal Sentence Encoder's ones, and in their constrastive2 run, they trained the model for 1,350 epochs rather than for 1,500 epochs.

Team \textbf{IPPAN} first extracted various features about the claims, including bag-of-words $n$-grams, word2vec vector representations~\cite{mikolov-yih-zweig:2013:NAACL-HLT}, named entity types, part-of-speech tags, sentiment scores, and features from statistical analysis of the sentences~\cite{checkthat19:Gasior}. Then, they used these features in an L1-regularized logistic regression to predict the check-worthiness of the sentences.

Team \textbf{Terrier} represented the sentences using bag of words and named entities~\cite{checkthat19:Su}. They used co-reference resolution to substitute the pronouns by the referring entity/person name. They further computed entity similarity~\cite{zhu2016computing} and entity relatedness~\cite{zhu2015sematch}. For prediction, they used an SVM classifier.

Team \textbf{UAICS} used a Na\"ive Bayes classifier with bag-of-words features~\cite{checkthat19:Coca}. In their contrastive submissions, they used other models, e.g., logistic regression.

Team \textbf{Factify} used the pre-trained ULMFiT model~\cite{howard2018universal} and fine-tuned it on the training set. They further over-sampled the minority class by replacing words randomly with similar words based on word2vec similarity. They also used data augmentation based on back-translation, where each sentence was translated to French, Arabic and Japanese and then back to English.

Team \textbf{JUNLP} extracted various features, including syntactic $n$-grams, sentiment polarity, text subjectivity, and LIX readability score, and used them to train a logistic regression classifier with high recall~\cite{checkthat19:Dhar}. Then, they trained an LSTM model fed with word representations from GloVe and part-of-speech tags. The sentence representations from the LSTM model were concatenated with the extracted features and used for prediction by a fully connected layer, which had high precision. Finally, they averaged the posterior probabilities from both models in order to come up with the final check-worthiness score for the  sentence.

\begin{table}[t]
\centering
\caption{Overview of the approaches to Task~1: check-worthiness. The learning model and the representations for the best system~\cite{checkthat19:Hansen} are highlighted.}
\label{tab:overviewt1}
\begin{tabular}{@{}c @{\hspace{1mm}} c@{}}
\begin{tabular}{l @{\hspace{-0mm}} c @{\hspace{-0mm}} c @{\hspace{-0mm}} c @{\hspace{-0mm}} c @{\hspace{-0mm}} c @{\hspace{-0mm}} c@{\hspace{-0mm}} c@{\hspace{-0mm}} c}
\toprule
\bf Learning Models     & \cite{checkthat19:Altun} & \cite{checkthat19:Coca}    & \cite{checkthat19:Dhar}   & \cite{checkthat19:Favano} & \cite{checkthat19:Gasior} & \cite{checkthat19:Hansen} & \cite{checkthat19:Mohtaj} & \cite{checkthat19:Su}  \\ \midrule

Neural Networks     \\
\,\,\, LSTM             &                          &                            & \faCheckSquare            &                           &                           & \red{\faCheckSquare}      &                           &                         \\
\,\,\, Feed-forward     &                          &                            &                           & \faCheckSquare            &                           &                           &                           &                         \\
SVM                     &                          &                            &                           &                           &                           &                           &                           &  \faCheckSquare         \\
Na\"ive Bayes             &                          & \faCheckSquare             &                           &                           &                           &                           &                           &                         \\
Logistic regressor      &                          &                            & \faCheckSquare            &                           &                           &                           &                           &                         \\
Regression trees & \faCheckSquare &                           &                           &                                                    &                           &                           &                         \\\hline
\\
\bf Teams		&	\\

\cite{checkthat19:Altun} TOBB ETU   & \multicolumn{8}{l}{\cite{checkthat19:Mohtaj} é proibido cochilar}	\\
\cite{checkthat19:Coca}  UAICS      & \multicolumn{8}{l}{\cite{checkthat19:Su}  Terrier}\\	
\cite{checkthat19:Dhar} JUNLP       & \multicolumn{8}{l}{[--] IIT (ISM) Dhanbad}\\	
\cite{checkthat19:Favano} TheEarthIsFlat    & \multicolumn{8}{l}{[--] Factify}\\	
\cite{checkthat19:Gasior}  IPIPAN       & \multicolumn{8}{l}{[--] nlpir01}\\
\cite{checkthat19:Hansen} Copenhagen        \\
\\
\bottomrule
\end{tabular}
&
\begin{tabular}{l @{\hspace{0mm}} c @{\hspace{-0mm}} c @{\hspace{-0mm}} c @{\hspace{-0mm}} c @{\hspace{-0mm}} c @{\hspace{-0mm}} c @{\hspace{-0mm}} c @{\hspace{-0mm}} c}
\toprule
\bf Represent.	    & \cite{checkthat19:Altun} & \cite{checkthat19:Coca}    & \cite{checkthat19:Dhar}   & \cite{checkthat19:Favano} & \cite{checkthat19:Gasior} & \cite{checkthat19:Hansen} & \cite{checkthat19:Mohtaj} & \cite{checkthat19:Su}  \\ \midrule

Embeddings              &                          &                            &                           &                           &                           &                           &                           &                         \\
\,\,\,PoS               &                          &                            & \faCheckSquare            &                           &                           &                           &                           &                         \\
\,\,\,word              &                          &                            & \faCheckSquare            &                           &  \faCheckSquare           & \red{\faCheckSquare}      &                           &                         \\
\,\,\,syntactic dep.    &                          &                            &                           &                           &                           & \red{\faCheckSquare}      &                           &                         \\
\,\,\,SUSE              &                          &                            &                           & \faCheckSquare            &                           &                           &                           &                         \\
Bag of \ldots           \\
\,\,\, words            &			               & \faCheckSquare   			&                			&                	        &  \faCheckSquare           &                	        &                			& \faCheckSquare	        \\                        
\,\,\, $n$-grams        & \faCheckSquare           &                            &                           &                           &                           &                           &                           &  \\
\,\,\, NEs   & \faCheckSquare           &                            &                           &                           & \faCheckSquare            &                           &                           & \faCheckSquare           \\
\,\,\, PoS   & \faCheckSquare           &                            &                           &                           & \faCheckSquare            &                           &                           &                           \\
Readability      &                          &                            &                           & \faCheckSquare            &                           &                           &                           &                           \\
Synt. $n$-grams     &                          &                            &                           & \faCheckSquare            &                           &                           &                           &                           \\
Sentiment               &                          &                            &                           & \faCheckSquare            & \faCheckSquare            &                           &                           &                           \\
Subjectivity            &                          &                            &                           & \faCheckSquare            &                           &                           &                           &                           \\
Sent. context    &                          &                            &                           & \faCheckSquare            &                           &                           &                           &                         \\
Topics                  & \faCheckSquare           &                            &                           &                           &                           &                           &                           &  \\\bottomrule

\end{tabular}
\end{tabular}
\end{table}

Team \textbf{nlpir01} extracted features such as tf-idf word vectors, tf-idf PoS vectors, word, character, and PoS tag counts. Then, they used these features in a multi-layer perceptron regressor with two hidden layers, each of size 2,000. For their contrastive1 run, they oversampled the minority class, and for their contrastive2 run, they reduced the number of units in each layer to 480.

Team \textbf{TOBB ETU} used linguistic features such as named entities, topics extracted with IBM Watson’s NLP tools, PoS tags, bigram counts and indicators of the type of sentence to train a multiple additive regression tree~\cite{checkthat19:Altun}. They further decreased the ranks of some sentences using hand-crafted rules. In their contrastive1 run, they added the speaker as a feature, while in their contrastive2 run they used logistic regression.

Team \textbf{IIT (ISM) Dhanbad} trained an LSTM-based recurrent neural network. They fed the network with word2vec embeddings and features extracted from constituency parse trees as well as features based on named entities and sentiment analysis.

Team \textbf{é proibido cochilar} trained an SVM model on bag-of-words representations of the sentences, after performing co-reference resolution and removing all digits~\cite{checkthat19:Mohtaj}. They further used an additional corpus of labeled claims, which they extracted from fact-checking websites, aiming at having a more balanced training corpus and potentially better generalizations.\footnote{Their claim crawling tool: \url{http://github.comx/vwoloszyn/fake\_news\_extractor}}

\subsection{Evaluation}

\subsubsection{Evaluation Measures}

Task~1 is shaped as an information retrieval problem, in which check-worthy instances should be ranked at the top of the list. Hence, we use mean average precision (MAP) as the official evaluation measure, which is defined as follows:

\begin{equation}
 MAP = \frac{\sum_{d=1}^D AveP(d)}{D}
\end{equation}
where $d\in D$ is one of the debates/speeches, and $AveP$ is the average precision, which in turn is defined as follows:

\begin{equation}
 AveP = \frac{\sum_{k=1}^K (P(k)\times \delta (k))}{\mbox{\# check-worthy claims}  }
\end{equation}
where $P(k)$ refers to the value of precision at rank $k$ and $\delta(k)=1$ iff the claim at that position is actually check-worthy. 

As in the 2018 edition of the task~\cite{clef2018checkthat:task1}, following~\cite{gencheva-EtAl:2017:RANLP} we further report some other measures: (\emph{i})~mean reciprocal rank (MRR), (\emph{ii})~mean R-Precision (MR-P), and (\emph{iii})~mean precision@$k$ (P@$k$). Here \emph{mean} refers to macro-averaging over the testing debates/speeches.

\begin{table}[t]
\caption{Results for Task~1: Check-worthiness. The results for the primary submission appear next to the team's identifier, followed by the contrastive submissions (if any). The subscript numbers indicate the rank of each primary submission with respect to the corresponding evaluation measure.}
\label{tab:results-t1}
\centering
\resizebox{\textwidth}{!}{%
\begin{tabular}{lllllllllll}
\toprule
        & \bf Team                & \bf MAP            &  RR            &  R-P           & P@1        &  P@3        &  P@5        &  P@10          &  P@20          &  P@50   \\
\midrule
\cite{checkthat19:Hansen} &\bf Copenhagen          & \bf \underline{.1660$_1$}    & .4176$_3$    & .1387$_4$    & .2857$_2$ &\bf .2381$_1$ &\bf .2571$_1$ & .2286$_2$    & .1571$_2$    & .1229$_2$ \\
\scriptsize        &\scriptsize \,\,\,\,\,contr.-1  &\scriptsize .1496         & \scriptsize .3098        & \scriptsize .1297        &\scriptsize .1429     &\scriptsize .2381     & \scriptsize .2000     & \scriptsize .2000        &\scriptsize  .1429        &\scriptsize  .1143 \\
\scriptsize        &\scriptsize \,\,\,\,\,contr.-2  & \scriptsize .1580         &\scriptsize .2740        &\scriptsize .1622        &\scriptsize .1429     &\scriptsize .1905     &\scriptsize .2286     &\scriptsize .2429        &\scriptsize .1786        &\scriptsize .1200 \\
\midrule
\cite{checkthat19:Favano} &\bf TheEarthIsFlat      & .1597$_2$     & .1953$_{11}$ &\bf .2052$_1$    & .0000$_4$ & .0952$_3$ & .2286$_2$ & .2143$_3$    &\bf .1857$_1$    &\bf .1457$_1$ \\
\scriptsize        &\scriptsize \,\,\,\,\,contr.-1  &\scriptsize .1453         &\scriptsize .3158        &\scriptsize .1101        &\scriptsize .2857     &\scriptsize .2381     &\scriptsize .1429     & \scriptsize.1429        &\scriptsize .1357        &\scriptsize .1171 \\
\scriptsize        &\scriptsize \,\,\,\,\,contr.-2  &\scriptsize .1821         &\scriptsize .4187        &\scriptsize .1937        &\scriptsize .2857     &\scriptsize .2381     &\scriptsize .2286     &\scriptsize .2286        &\scriptsize .2143        &\scriptsize .1400 \\
\midrule
\cite{checkthat19:Gasior} &\bf IPIPAN              & .1332$_3$     & .2864$_6$    & .1481$_2$    & .1429$_3$ & .0952$_3$ & .1429$_5$ & .1714$_5$    & .1500$_3$    & .1171$_3$ \\
\midrule
\cite{checkthat19:Su} & \bf Terrier             & .1263$_4$     & .3253$_5$    & .1088$_8$    & .2857$_2$ &\bf .2381$_1$ & .2000$_3$ & .2000$_4$    & .1286$_6$    & .0914$_7$ \\
\midrule
\cite{checkthat19:Coca} &\bf UAICS               & .1234$_5$     &\bf\bf .4650$_1$    & .1460$_3$    &\bf .4286$_1$ &\bf .2381$_1$ & .2286$_2$ &\bf .2429$_1$    & .1429$_4$    & .0943$_6$ \\
\scriptsize        &\scriptsize \,\,\,\,\,contr.-1  &\scriptsize .0649         &\scriptsize .2817        & \scriptsize.0655        &\scriptsize .1429     &\scriptsize .2381     &\scriptsize .1429     &\scriptsize .1143        & \scriptsize .0786        &\scriptsize .0343 \\
        &\scriptsize \,\,\,\,\,contr.-2  &\scriptsize .0726         &\scriptsize .4492        &\scriptsize .0547        &\scriptsize .4286     &\scriptsize .2857     &\scriptsize .1714     &\scriptsize .1143        &\scriptsize .0643        &\scriptsize .0257 \\
\midrule
 & \bf Factify             & .1210$_6$     & .2285$_8$    & .1292$_5$    & .1429$_3$ & .0952$_3$ & .1143$_6$ & .1429$_6$    & .1429$_4$    & .1086$_4$ \\
\midrule
\cite{checkthat19:Dhar} &\bf JUNLP               & .1162$_7$     & .4419$_2$    & .1128$_7$    & .2857$_2$ & .1905$_2$ & .1714$_4$ & .1714$_5$    & .1286$_6$    & .1000$_5$ \\
\scriptsize        &\scriptsize \,\,\,\,\,contr.-1  &\scriptsize .0976         &\scriptsize .3054        &\scriptsize .0814        &\scriptsize .1429     &\scriptsize .2381     &\scriptsize .1429     & \scriptsize .0857        & \scriptsize .0786        &\scriptsize .0771 \\
        &\scriptsize \,\,\,\,\,contr.-2  &\scriptsize .1226         &\scriptsize .4465        &\scriptsize .1357        &\scriptsize .2857     &\scriptsize .2381     &\scriptsize .2000     &\scriptsize .1571        &\scriptsize .1286        &\scriptsize .0886 \\
\midrule
&\bf nlpir01             & .1000$_8$     & .2840$_7$    & .1063$_9$    & .1429$_3$ &\bf .2381$_1$ & .1714$_4$ & .1000$_8$    & .1214$_7$    & .0943$_6$ \\
        &\scriptsize \,\,\,\,\,contr.-1  &\scriptsize .0966         &\scriptsize .3797        &\scriptsize .0849        &\scriptsize .2857     &\scriptsize .1905     &\scriptsize .2286     &\scriptsize .1429        &\scriptsize .1071        &\scriptsize .0886 \\
        &\scriptsize \,\,\,\,\,contr.-2  &\scriptsize .0965         &\scriptsize .3391        &\scriptsize .1129        &\scriptsize .1429     &\scriptsize .2381     &\scriptsize .2286     &\scriptsize .1571        &\scriptsize .1286        &\scriptsize .0943 \\
\midrule
\cite{checkthat19:Altun} &\bf TOBB ETU            & .0884$_9$     & .2028$_{10}$ & .1150$_6$    & .0000$_4$ & .0952$_3$ & .1429$_5$ & .1286$_7$    & .1357$_5$    & .0829$_8$ \\
        &\scriptsize \,\,\,\,\,contr.-1  &\scriptsize .0898         &\scriptsize .2013        &\scriptsize .1150        &\scriptsize .0000     &\scriptsize .1429     &\scriptsize .1143     & \scriptsize.1286        &\scriptsize .1429        &\scriptsize .0829 \\
        &\scriptsize \,\,\,\,\,contr.-2  &\scriptsize .0913         &\scriptsize .3427        &\scriptsize .1007        &\scriptsize .1429     &\scriptsize .1429     &\scriptsize .1143     &\scriptsize .0714        &\scriptsize .1214        &\scriptsize .0829 \\
\midrule
& \bf IIT (ISM) Dhanbad   & .0835$_{10}$  & .2238$_9$    & .0714$_{11}$ & .0000$_4$ & .1905$_2$ & .1143$_6$ & .0857$_9$    & .0857$_9$    & .0771$_9$ \\
\midrule
\cite{checkthat19:Mohtaj} &\bf é proibido cochilar & .0796$_{11}$& .3514$_4$    & .0886$_{10}$ & .1429$_3$ &\bf .2381$_1$ & .1429$_5$ & .1286$_7$    & .1071$_8$    & .0714$_{10}$ \\
        &\scriptsize \,\,\,\,\,contr.-1  &\scriptsize .1357         &\scriptsize .5414        &\scriptsize .1595        &\scriptsize .4286     &\scriptsize .2381     &\scriptsize .2571     & \scriptsize .2714        &\scriptsize .1643        & \scriptsize .1200 \\
\bottomrule
\end{tabular}}
\end{table}

\subsubsection{Results}

The participants were allowed to submit one primary and up to two contrastive runs in order to test variations of their primary models or entirely different alternative models. Only the primary runs were considered for the official ranking. A total of eleven teams submitted 21 runs. Table~\ref{tab:results-t1} shows the results.

The best-performing system was the one by team \textbf{Copenhagen}. They achieved a strong MAP score using a ranking loss based on contrastive sampling. Indeed, this is the only team that modeled the task as a ranking one and the decrease in the performance without the ranking loss (see their contrastive1 run) shows the importance of using this loss.

Two teams made use of external datasets: team \textbf{Copenhagen} used a weakly supervised dataset for pretraining, and team \textbf{é proibido cochilar} included claims scraped from several fact-checking Web sites.
In order to address the class imbalance in the training dataset, team \textbf{nlpir01} used oversampling in their contrastive1 run, but could not gain any improvements. Oversampling and augmenting with additional data points did not help team \textbf{é proibido cochilar} either.

Many systems used pretrained sentence or word embedding models. Team \textbf{TheEarthIsFlat}, which is the second-best performing system, used the Standard Universal Sentence Embeddings, which performed well on the task. The best MAP score overall was obtained by the second contrastive run by this team: the only difference with respect to their primary submission was the number of training epochs. Some teams also used fine-tuning, e.g.,~team \textbf{Factify} fine-tuned the ULMFiT model on the training dataset.


\section{Overview of Task 2: Evidence and Factuality}
\label{sec:task2}

\noindent
Task 2 focuses on building tools to verify the factuality of a given check-worthy claim. This is the first-ever version of this task, and we run it in Arabic.\footnote{In 2018, we had a different fact-checking task, where no retrieved Web pages were provided~\cite{clef2018checkthat:task2}.} The task is formally defined as follows: 
\begin{center}
\parbox[c]{0.85\textwidth}{%
\textit{Given a check-worthy claim $c$ associated with a set of Web pages $P$ (that constitute the retrieved results of Web search in response to a search query that represents the claim), identify which of the Web pages (and passages $A$ of those Web pages) can be useful for assisting a human in fact-checking the claim. Finally, determine the factuality of the claim according to the supporting information in the useful pages and passages.
}}
\end{center}

As Figure~\ref{fig:pipelinetask2} shows, the task is divided into four subtasks that target different aspects of the problem:

\begin{figure}[t]
\centering
\includegraphics[width=1.0\columnwidth]{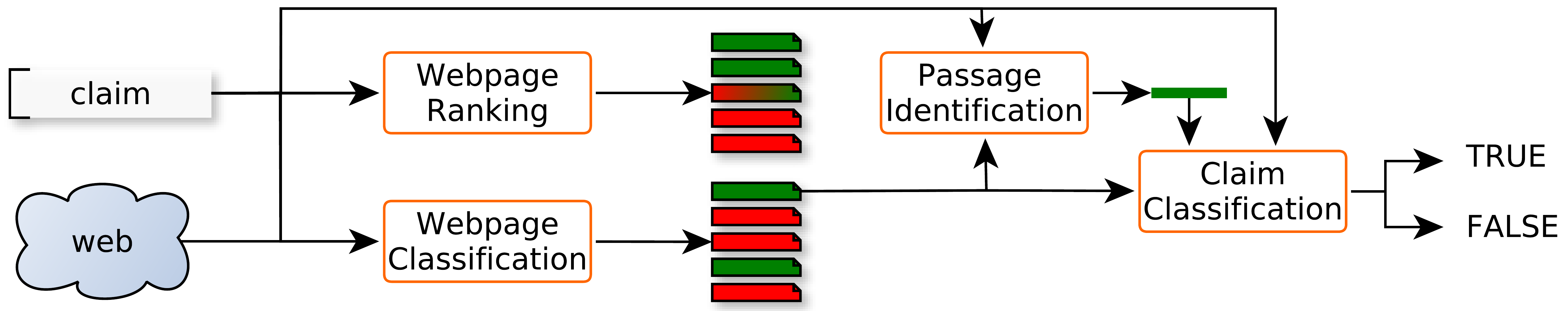}
\vspace*{-1mm}
\caption{A zoom into the four subtasks in Task 2.}
\label{fig:pipelinetask2}
\end{figure}

\begin{description}[noitemsep,topsep=0pt]
\itemsep0em
\item[Subtask A, Rerank search results:] \emph{Rank the Web pages $P$ based on how useful they are for verifying the target claim}. The systems are required to produce a score for each page, based on which the pages would be ranked. See the definition of ``useful'' pages below.
\item[Subtask B, Classify search results:] \emph{Classify each of the Web pages $P$ as ``very useful for verification'', ``useful'', ``not useful'', or ``not relevant.''} A page is considered \emph{very useful} for verification if it is \emph{relevant} with respect to the claim (i.e.,~on-topic and discussing the claim) and it \emph{provides sufficient evidence} to verify the veracity of the claim, such that there is no need for another document to be considered for verifying this claim. A page is \emph{useful} for verification if it is relevant to the claim and provides some valid evidence, but it is \emph{not solely sufficient} to determine the claim's veracity on its own. The evidence can be a source, some statistics, a quote, etc. However, a particular piece of evidence is considered not valid if the source cannot be verified or is ambiguous (e.g.,~expressing that ``experts say that\ldots'' without mentioning who those experts are), or it is just an opinion of a person/expert instead of an objective analysis. 

Notice that this is different from \emph{stance detection}, as a page might agree with a claim, but it might still lack evidence to verify it.
\item[Subtask C, Classify passages from useful/very useful pages:] \emph{Find passages within those Web pages that are \textbf{useful} for claim verification.} Again, notice that this is different from stance detection. 
\item[Subtask D, Verify the claim:] \emph{Classify the claim's factuality as ``true'' or ``false.''} 
The claim is considered true if it is accurate as stated (or there is sufficient reliable evidence supporting it), otherwise it is considered false. 

\end{description}

\begin{figure}[t]
\centering
\includegraphics[width=\textwidth]{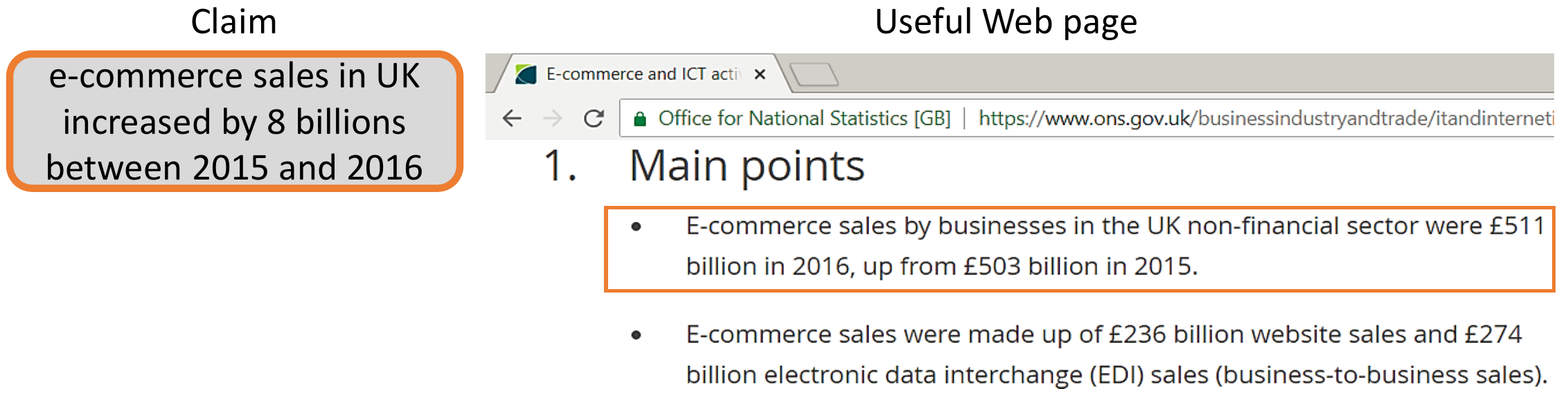}
\vspace*{-1mm}
\caption{A claim, a useful Web page, and a useful passage (in the box).}
\label{fig:extask2}
\end{figure}

Figure~\ref{fig:extask2} shows an example. For the sake of readability, the example is given in English, but this year the task was offered only in Arabic. The example shows a Web page that is considered useful for verifying the given claim, since it has evidence showing the claim to be true and as the page itself is an official United Kingdom page on national statistics. The useful passage in the page is the one reporting the supporting statistics.

\subsection{Dataset}
\textbf{Collecting Claims.} Subtasks A, B, and C are all new to the lab this year. 
As a result, we built a new evaluation dataset to support all subtasks ---the CT19-T2 corpus. We selected 69 claims from multiple sources including a pre-existing set of Arabic claims~\cite{baly2018integrating}, a survey in which we asked the public to provide examples of claims they have heard of, and some headlines from six Arabic news agencies that we rewrote into claims. The news agencies selected are well-known in the Arab world: Al Jazeera, BBC Arabic, CNN Arabic, Al Youm Al Sabea, Al Arabiya, and RT Arabic. We made sure the claims span different topical domains, e.g., health or sports, besides politics. Ten claims were released for training and the rest were used for testing.
\vspace{12pt}

\noindent
\textbf{Labelling Claims.} We acquired the veracity labels for the claims in two steps. First, two of the lab organizers labelled each of the 69 claims independently. Then, they met to resolve any disagreements, and thus reach consensus on the veracity labels for all claims. 
\vspace{12pt}

\noindent
\textbf{Labelling Pages and Passages.} For each claim, we formulated a query representing the claim, and we issued it against the Google search engine in order to extract the top 100 Web pages returned as a result. We used a language detection tool to filter out non-Arabic pages, and we eventually used the top-50 of the remaining pages. The labelling pipeline was carried out as follows:

\begin{enumerate}
    \item \textbf{Relevance}. We first identified relevant pages, since we assume that non-relevant pages cannot be useful for claim verification, and thus should be filtered out from any further labelling. In order to speedup the relevance labelling process, we hired two types of annotators: crowd-workers, through Amazon Mechanical Turk, and in-house annotators. Each page was labelled by \textit{three} annotators, and the majority label was used as the final page label.
    \item \textbf{Usefulness as a whole}. Relevant pages were then given to in-house annotators to be labelled for usefulness using a two-way classification scheme: \emph{useful} (including \emph{very useful}, but not distinguishing between the two) and \emph{not useful}. Similar to relevance labelling, each page was labelled by three annotators, and the final page label was the majority label. 
    \item \textbf{Useful vs.\ very useful}. One of the lab organizers went over the useful pages (from the previous step) and further classified them into \emph{useful} and \emph{very useful}. We opted for this design since we found through pilot studies that the annotators found it difficult to differentiate between \emph{useful} and \emph{very useful} pages. 
    \item \textbf{Splitting into passages}. We further manually split the \emph{useful} and the \emph{very useful} pages into passages, as we found that the automatic techniques for splitting pages into passages were not accurate enough.
    \item \textbf{Useful passages}. Finally, one of the lab organizers labelled each passage for usefulness. Due to time constraints, we could not split the pages and label the resulting passages for all the claims in the \emph{testing set}. Thus, we only release labels for passages of pages corresponding to 33 out of the 59 testing claims. Note that this only affects subtask C.
\end{enumerate}

Table~\ref{tab:task2datasets} summarizes the statistics about the training and the test data for Task~2. Note that the passages in the test set are for 33 claims only (see above).

\begin{table}[t]
\centering
\caption{Statistics about the CT19-T2 corpus for Task 2.\label{tab:task2datasets}}
\begin{tabular}{l@{ }@{ }@{ }@{ }@{ }rr@{ }@{ }@{ }@{ }@{ }@{ }rr@{ }@{ }@{ }@{ }@{ }@{ }rr}
\toprule
  & \multicolumn{2}{c}{\bf Claims} & \multicolumn{2}{c}{\bf Pages} & \multicolumn{2}{c}{\bf Passages} \\ {\bf Set} & {\bf Total} & {\bf True} & {\bf Total} & {\bf Useful} & {\bf Total} & {\bf Useful} \\
\midrule
{\bf Training} &         10 &          5 &        395 &         32 &        167 &         54 \\
{\bf Test} &         59 &         30 &       2,641 &        575 &       1,722 &        578 \\
\bottomrule
\end{tabular}  
\end{table}

\subsection{Subtask A}

\textbf{Runs}
Three teams participated in this subtask submitting a total of seven runs~\cite{checkthat19:Favano,checkthat19:Haouari,checkthat19:Touahri}. There were two kinds of approaches. In the first kind, token-level BERT embeddings were used with text classification to rank pages~\cite{checkthat19:Favano}. In the second kind, runs used a learning-to-rank model based on different classifiers, including Na\"ive Bayes and Random Forest, with a variety of features for ranking~\cite{checkthat19:Haouari}. In one run, external data was used to train the text classifier~\cite{checkthat19:Favano}, while all other runs represent systems trained on the provided labelled data only.
\vspace{6pt}

\noindent
\textbf{Evaluation Measures}
Subtask A was modelled as a ranking problem, in which \emph{very useful} and \emph{useful} pages should be ranked at the top. Since this is a graded usefulness problem, we evaluate it using the mean of Normalized Discounted Cumulative Gain (nDCG)~\cite{jarvelin2002cumulated,manning2010}. In particular, we consider nDCG@10 (i.e.,~nDCG computed at cutoff 10) as the official evaluation measure for this subtask, but we report nDCG at cutoffs 5, 15, and 20 as well. 

We also report precision at cutoffs 5, 10, 15, and 20, in addition to Mean Average Precision (MAP). 
For precision-based measures, we consolidate the labels into two labels instead of four: we combined the \emph{very useful} and the \emph{useful} pages under the \emph{useful} label, and we considered the rest as \emph{not useful}. In all measures, we used macro-averaging over the testing claims.

\noindent
\textbf{Results.} Table~\ref{tab:t2:ta} shows the results for all seven runs. It also includes the results of a simple baseline: the original ranking in the search result list. We can see that the baseline surprisingly performs the best. This is due to the fact that in our definition of usefulness, useful pages must be relevant, and Google, as an effective search engine, has managed to rank relevant pages (and consequently, many of the \emph{useful} pages) first. This result indicates that the task of ranking pages by usefulness is not easy and systems need to be further developed in order to differentiate between relevance and usefulness, while also benefiting from the relevance-based rank of a page.

\begin{table}[t]
\centering
\caption{\label{tab:t2:ta} Results for Subtask 2.A, ordered by nDCG@10 score. The runs that used external data are marked with a *.}
\begin{tabular}{lc@{ }@{ }@{ }c@{ }@{ }@{ }c@{ }@{ }@{ }c@{ }@{ }@{ }c}
\toprule
 \bf Team      & \bf Run& \bf nDCG@5 & \bf nDCG@10 & {\bf nDCG@15} & {\bf nDCG@20} \\
\midrule
\textbf{Baseline}   &   & \bf 0.52 &  \bf 0.55 &       \textbf{0.58} &       \textbf{0.61} \\
bigIR              & 1 &  0.47 &       0.50 &       0.54 &       0.55 \\
bigIR              & 3 &  0.41 &       0.47 &       0.50 &       0.52 \\
EvolutionTeam       & 1 &  0.40 &       0.45 &       0.48 &       0.51 \\
bigIR              & 4 &  0.39 &       0.45 &       0.48 &       0.51 \\
bigIR              & 2 &  0.38 &       0.41 &       0.45 &       0.47 \\
TheEarthIsFlat2A    & 1 &  0.08 &       0.10 &       0.12 &       0.14 \\
TheEarthIsFlat2A*   & 2 &  0.05 &       0.07 &       0.10 &       0.12 \\
\bottomrule
\end{tabular}
\end{table}

\subsection{Subtask B}

\noindent
\textbf{Runs}.
Four teams participated in this subtask, submitting a total of eight runs~\cite{checkthat19:Favano,checkthat19:Ghanem,checkthat19:Haouari,checkthat19:Touahri}. All runs used supervised text classification models, such as Random Forest and Gradient Boosting~\cite{checkthat19:Haouari}. In terms of representation, two teams opted for using embedding-based language representation, with one team using word embeddings~\cite{checkthat19:Ghanem}, and the other one opting for BERT-based token-level embeddings for all their runs~\cite{checkthat19:Favano}. In one run, external data was used to train the model~\cite{checkthat19:Favano}, while all the remaining runs were trained on the provided labelled training data only.
\vspace{6pt}

\noindent
\textbf{Evaluation Measures}.
Similar to Subtask A, Subtask B also aims at identifying useful pages for claim verification, but it is modeled as a \emph{classification} problem, while Subtask A was a ranking problem. Thus, for evaluation we use standard evaluation measures for text classification: Precision, Recall, $F_1$, and Accuracy, with $F_1$ being the official score for the task. 
\vspace{6pt}

\noindent
\textbf{Results}. Table~\ref{tab:t2} shows the results. 
Table~\ref{tab:t2:tb1} reports the results for 2-way classification ---\emph{useful/very useful} vs. \emph{not useful/not relevant}---, reporting results for predicting the \emph{useful} class. 
Table~\ref{tab:t2:tb2} shows the results for 4-way classification ---\emph{very useful} vs. \emph{useful} vs. \emph{not useful} vs. \emph{not relevant}---, reporting macro-averaged scores over the four classes, for each of the evaluation measures. 

We also included a baseline, which is based on the original ranking in the search results list. The baseline assumes the top-50\% of the results to be \emph{useful} and the rest \emph{not useful} for the 2-way classification. For the 4-way classification, the baseline assumes the top-25\% to be \emph{very useful}, the next 25\% to be \emph{useful}, the third 25\% to be \emph{not useful}, and the rest to be \emph{not relevant}.

\begin{table}[t]
\caption{\label{tab:t2} Results for Subtask 2.B for 2-way and 4-way classification. 
The runs are ranked by $F_1$ score. Runs tagged with a * used external data.}
\begin{subtable}{.5\textwidth}
\caption{2-way classification \label{tab:t2:tb1}}

\begin{tabular}{l @{\hspace{-2mm}}ccccc}
\toprule
 {\bf Team}             & \bf Run  & \bf F$_1$  & {\bf P}    & {\bf R}    & {\bf Acc} \\
\midrule
\bf Baseline            &  & \bf 0.42 & \bf 0.30 & \bf 0.72 & \bf 0.57 \\
UPV-UMA         & 1 &      0.38 &       0.26 &       0.73  &       0.49 \\
bigIR          & 1 &      0.08 &       0.40 &       0.04  &       0.78 \\
bigIR          & 3 &      0.07 &       0.39 &       0.04  &       0.78 \\
bigIR          & 4 &      0.07 &       0.57 &       0.04  &       0.78 \\

bigIR          & 2 &      0.04 &       0.22 &       0.02  &       0.77 \\
TheEarthIsFlat  & 1 &      0.00 &       0.00 &       0.00  &       0.78 \\
TheEarthIsFlat* & 2 &      0.00 &       0.00 &       0.00  &       0.78 \\

EvolutionTeam   & 1  &      0.00 &       0.00 &       0.00  &       0.78 \\

\bottomrule
\end{tabular}  
\end{subtable}
\begin{subtable}{.5\textwidth}
\caption{4-way classification \label{tab:t2:tb2}}
\begin{tabular}{l @{\hspace{-2mm}}ccccc}
\toprule
\bf Team  & \bf Run   & {\bf F$_1$} & {\bf P} & {\bf R} &    {\bf Acc} \\
\midrule
TheEarthIsFlat  & 1   &    0.31 &   0.28 &       0.36 &              0.59 \\
bigIR          & 3   &     0.31 &    0.37 &       0.33 &            0.58 \\
TheEarthIsFlat* & 2   &    0.30 &   0.27 &       0.35 &              0.60 \\
bigIR          & 4   &    0.30 &   0.41 &       0.32 &              0.57 \\
EvolutionTeam   & 1   &    0.29 &   0.26 &       0.33 &              0.58 \\
{\bf Baseline}  &     & {\bf 0.28} &{\bf 0.32} & {\bf 0.32} &  {\bf 0.30} \\
UPV-UMA         & 1   &    0.23 &   0.30 &       0.29 &              0.24 \\
bigIR          & 1   &    0.16 &  0.25 &       0.23 &              0.26 \\
bigIR          & 2   &    0.16 &   0.25 &       0.22 &              0.25 \\
\bottomrule
\end{tabular}    
\end{subtable}

\end{table}

Table~\ref{tab:t2:tb1} shows that almost all systems struggled to retrieve any \emph{useful} pages at all. Team UPV-UMA is the only one that managed to achieve high recall. 
This is probably due to the \emph{useful} class being under-represented in the training dataset, while being much more frequent in the test dataset: we can see in Table~\ref{tab:task2datasets} that it covers just 8\% of the training examples, but 22\% of the testing ones.
 Training the models with a limited number of \emph{useful} pages might have caused them to learn to underpedict this class.
Similar to Subtask A, the simple baseline that assumes the top-ranked pages to be more useful is most effective. This again can be due to the correlation between usefulness and relevance.

%
%
%
%
%
%

Comparing the results in Table~\ref{tab:t2:tb1} to those in Table~\ref{tab:t2:tb2}, we notice a very different performance ranking; runs that had the worst performance at finding \emph{useful} pages, are actually among the best runs in the 4-way classification. These runs were able to effectively detect the \emph{not relevant} and \emph{not useful} pages as compared to \emph{useful} ones. The baseline, which was effective at identifying \emph{useful} pages, is not as effective at identifying pages in the other classes. This might indicate that \emph{not useful} and \emph{not relevant} pages are not always at the bottom of the ranked list as this baseline assumes, which sheds some light on the importance of usefulness estimation to aid fact-checking. One additional factor that might have caused such varied ranking of runs is our own observation on the difficulty and subjectivity of differentiating between \emph{useful} and \emph{very useful} pages. At annotation time, we observed that annotators and even lab organizers were not able to easily distinguish between these two types of pages.

\subsection{Subtask C}

\noindent
\textbf{Runs}.
Two teams participated in this task~\cite{checkthat19:Favano,checkthat19:Haouari}, submitting a total of seven runs. One of the teams used text classifiers including Na\"ive Bayes and SVM with a variety of features such as bag-of-words and named entities~\cite{checkthat19:Haouari}. All runs also considered using the similarity between the claim and the passages as a feature in their models.  
\vspace{9pt}

\noindent
\textbf{Evaluation Measures} Subtask C aims at identifying useful passages for claim verification and we modelled it as a classification problem. As in typical classification problems, we evaluated it using Precision, Recall, $F_1$, and Accuracy, with $F_1$ being the official evaluation measure for the task.
\vspace{9pt}

\noindent
\textbf{Results} Table~\ref{tab:t2:tc} shows the evaluation results. The scores for precsion, recall and $F_1$ are calculated with respect to the positive class, i.e.,~\emph{useful}. The table also shows the evaluation results for a simple baseline that assumes the first passage in a page to be \emph{not useful}, the next two passages to be \emph{useful}, and the remaining passages to be \emph{not useful}. This baseline is motivated by our observation that \emph{useful} passages are typically located at the heart of the document following some introductory passage(s).
\vspace{3pt}

We can see that team TheEarthIsFlat managed to identify most of the \emph{useful} passages, thus achieving very high recall (0.94 for its run 1), while also having relatively similar precision to the other runs and the baseline. Further analysis of the performance of that system is needed in order to understand how it managed to achieve such a high recall. Note that in all the runs by the bigIR system, as well as in the baseline system, the precision and the recall are fairly balanced. We can see that the baseline performed almost as well as the four runs by bigIR. This indicates that considering the position of the passage in a page might be a useful feature when predicting the passage usefulness, and thus it should be considered when addressing that problem.

\begin{table}[t]
\centering
\caption{\label{tab:t2:tc} Performance of the models when predicting useful passages for Subtask 2.C. The runs are ranked by $F_1$.}
\begin{tabular}{l@{ }@{ }@{ }c@{ }@{ }@{ }c@{ }@{ }@{ }c@{ }@{ }@{ }c@{ }@{ }@{ }c}
\toprule
 {\bf Team}      & \bf Run            & \bf F$_1$ & \bf P & \bf R & \bf Acc \\
\midrule
TheEarthIsFlat2Cnoext & 1  &   0.56    & 0.40  & 0.94 &  0.51 \\
TheEarthIsFlat2Cnoext    & 2 &   0.55    & 0.41  & 0.87 &  0.53 \\
bigIR           & 2  &   0.40    & 0.39  & 0.42 &  0.58 \\
bigIR            & 1  &   0.39    & 0.38  & 0.41 &  0.58 \\
bigIR           & 4  &   0.37    & 0.37  & 0.38 &  0.57 \\
\bf Baseline              &   & {\bf 0.37} & {\bf 0.42} & {\bf 0.39} & {\bf 0.57} \\
bigIR           & 3  &   0.19    & 0.33  & 0.14 &  0.61 \\
\bottomrule
\end{tabular}  
\end{table}

\subsection{Subtask D}

The main aim of Task~2 was to study the effect of using identified \emph{useful} and \emph{very useful} pages for claim verification. Thus, we had two evaluation cycles for Subtask D. In the first cycle, the teams were asked to fact-check claims using the given Web pages, without knowing which of the Web pages were \emph{useful}/\emph{very useful}. In the second cycle, the usefulness labels were released in order to allow the systems to fact-check the claims when knowing which of the Web pages are \emph{useful}/\emph{very useful}.
\vspace{6pt}

\noindent
\textbf{Runs}
Two teams participated in cycle 1, submitting one run each~\cite{checkthat19:Haouari,checkthat19:Touahri}, but one of the runs was invalid, and thus there is only one official run. Cycle 2 attracted more participation: three teams with nine runs~\cite{checkthat19:Ghanem,checkthat19:Haouari,checkthat19:Touahri}.
Thus, we will focus our discussion on cycle 2. One team opted for using textual entailment with embedding-based representations for classification~\cite{checkthat19:Ghanem}. Another team used text classifiers such as Gradient Boosting and Random Forests~\cite{checkthat19:Haouari}. External data was used to train the textual entailment component of the system in four runs, whereas the remaining runs were trained on the provided data only.
\vspace{6pt}

\noindent
\textbf{Evaluation Measures} Subtask D aims at predicting a claim's veracity. It is a classification task, and thus we evaluate it using Precision, Recall, $F_1$, and Accuracy, with $F_1$ being the official measure for the task. 
\vspace{6pt}

\noindent
\textbf{Results} Table~\ref{tab:t2d} shows the results for cycles 1 and 2, where we macro-average precision, recall, and $F_1$ over the two classes. We show the results for a simple majority-class baseline, which all runs manage to beat for both cycles.
\vspace{6pt}

\begin{table}[t]
\caption{\label{tab:t2d} Results for Subtask 2.D for both cycles 1 and 2. The runs are ranked by $F_1$ score. The runs tagged with a * used external data.}
\begin{subtable}{0.47\textwidth}
\caption{Cycle 1, where the usefulness of the Web pages was unknown. \label{tab:t2:td1}}
\begin{tabular}{lc@{ }@{ }c@{ }@{ }c@{ }@{ }c}
\toprule
 {\bf Team}      & \bf F$_1$ & \bf P     & \bf R     & \bf Acc \\
\midrule
EvolutionTeam   & 0.48      & 0.55      &  0.53     &        0.53 \\
\bf Baseline  & \bf 0.34  & \bf 0.25  & \bf 0.50  &  \bf 0.51 \\
\bottomrule
\end{tabular} 

\begin{tabular}{l}
\\\\\\\\\\\\\\\\
\end{tabular}

\end{subtable}
\begin{subtable}{0.5\textwidth}
\caption{Cycle 2, where the the usefulness of the Web pages was known.}
\begin{tabular}{lc@{ }@{ }c@{ }@{ }c@{ }@{ }c@{ }@{ }c}
\toprule
\bf Team   & \bf Run & \bf F$_1$ & \bf P & {\bf R} &   {\bf Acc} \\
\midrule
UPV-UMA*        & 21 & 0.62     & 0.63 &       0.63 &              0.63 \\
 
UPV-UMA*        & 11 & 0.55     & 0.56 &       0.56 &              0.56 \\ 

UPV-UMA*        & 22 & 0.54     & 0.60 &       0.57 &              0.58 \\

bigIR          & 1  & 0.53     & 0.55 &       0.55 &              0.54 \\
bigIR          & 3  & 0.53     & 0.55 &       0.54 &              0.54 \\
bigIR          & 2  & 0.51     & 0.53 &       0.53 &              0.53 \\
bigIR          & 4  & 0.51     & 0.53 &       0.53 &              0.53 \\
UPV-UMA*        & 12 & 0.51     & 0.65 &       0.57 &              0.58 \\
EvolutionTeam   & 1  & 0.43     & 0.45 &       0.46 &              0.46 \\
{\bf Baseline}  &   &\bf 0.34     &\bf 0.25      &  \bf 0.50 &  \bf 0.51 \\
\bottomrule
\end{tabular}  
\end{subtable}
\end{table}

%
%
%
%
%
%
%
%

Due to the low participation in cycle~1, it is difficult to draw conclusions about whether providing systems with useful pages helps to improve their performance.

\section{Conclusion and Future Work}
\label{sec:conclusions}

We have presented an overview of the CLEF--2019 \ct \  Lab on Automatic Identification and Verification of Claims, which is the second edition of the lab. \ct \  proposed two complementary tasks. Task~1 asked the participating systems to predict which claims in a political debate should be prioritized for fact-checking. Task~2 was designed to aid a human who is fact-checking a claim. It asked the systems (A)~to rank Web pages with respect to a check-worthy claim based on their usefulness for fact-checking that claim, (B)~to classify the Web pages according to their degree of usefulness, (C)~to identify useful passages from these pages, and (D) to use the useful pages to predict a claim's factuality. As part of \ct, we release datasets in English (derived from fact-checking sources) and Arabic in order to enable further research in check-worthiness estimation and in automatic claim verification. 

A total of 14 teams participated in the lab (compared to 9 in 2018) submitting a total of 57 runs. The evaluation results show that the most successful approaches to Task 1 used various neural networks and logistic regression. As for Task 2, learning-to-rank was used by the highest-scoring runs for subtask A, while different classifiers were used in the other subtasks.

Regarding the task of selecting check-worthy claims, we consider expanding the dataset with more annotations. This will pave the way for the development of various neural architectures and it will also likely boost the accuracy of the final systems. We also plan to include more sources of annotations. As noted in \cite{gencheva-EtAl:2017:RANLP}, the agreement between different media sources on the task was low, meaning that there is a certain bias in the selection of the claims for each of the media outlets. Aggregating the annotations from multiple sources would potentially decrease this selection bias. 

Although one of the aims of the lab was to study the effect of using \emph{useful} pages for claim verification, the low participation in the first cycle of Subtask D of Task 2 has hindered carrying such a study. In the future, we plan to setup this subtask, so that teams must participate in both cycles in order for their runs to be considered valid. We also plan to extend the dataset for Task 2 to include claims in at least one other language than Arabic.

\section*{Acknowledgments}
This work was made possible in part by NPRP grant\# NPRP 7-1330-2-483 from the Qatar National Research Fund (a member of Qatar Foundation). The statements made herein are solely the responsibility of the authors. 

This research is also part of the Tanbih project,\footnote{\url{http://tanbih.qcri.org/}} which aims to limit the effect of ``fake news'', propaganda and media bias by making users aware of what they are reading. The project is developed in collaboration between the Qatar Computing Research Institute (QCRI), HBKU and the MIT Computer Science and Artificial Intelligence Laboratory (CSAIL).

\bibliographystyle{splncs04}
\bibliography{ref}

\begin{thebibliography}{10}
\providecommand{\url}[1]{\texttt{#1}}
\providecommand{\urlprefix}{URL }
\providecommand{\doi}[1]{https://doi.org/#1}

\bibitem{checkthat19:Altun}
Altun, B., Kutlu, M.: {TOBB-ETU} at {CLEF} 2019: Prioritizing claims based on
  check-worthiness. In: CLEF 2019 Working Notes. Working Notes of CLEF 2019 -
  Conference and Labs of the Evaluation Forum. {CEUR} Workshop Proceedings,
  CEUR-WS.org, Lugano, Switzerland (2019)

\bibitem{clef2018checkthat:task1}
Atanasova, P., M\`{a}rquez, L., Barr\'{o}n-Cede\~{n}o, A., Elsayed, T.,
  Suwaileh, R., Zaghouani, W., Kyuchukov, S., Da~San~Martino, G., Nakov, P.:
  Overview of the {CLEF-2018 CheckThat! Lab} on automatic identification and
  verification of political claims, {T}ask 1: Check-worthiness. In: Cappellato,
  L., Ferro, N., Nie, J.Y., Soulier, L. (eds.) CLEF 2018 Working Notes. Working
  Notes of CLEF 2018 - Conference and Labs of the Evaluation Forum. {CEUR}
  Workshop Proceedings, CEUR-WS.org, Avignon, France (2018)

\bibitem{Atanasova:2019:AFU:3331015.3297722}
Atanasova, P., Nakov, P., M\`{a}rquez, L., Barr\'{o}n-Cede\~{n}o, A.,
  Karadzhov, G., Mihaylova, T., Mohtarami, M., Glass, J.: Automatic
  fact-checking using context and discourse information. J. Data and
  Information Quality  \textbf{11}(3),  12:1--12:27 (May 2019)

\bibitem{ba2016vera}
Ba, M.L., Berti-Equille, L., Shah, K., Hammady, H.M.: {VERA}: A platform for
  veracity estimation over web data. In: Proceedings of the 25th International
  Conference Companion on World Wide Web. pp. 159--162. WWW~'16 (2016)

\bibitem{baly2018integrating}
Baly, R., Mohtarami, M., Glass, J., M\`{a}rquez, L., Moschitti, A., Nakov, P.:
  Integrating stance detection and fact checking in a unified corpus. In:
  Proceedings of the 2018 Conference of the North American Chapter of the
  Association for Computational Linguistics: Human Language Technologies. pp.
  21--27. NAACL-HLT~'18, New Orleans, Louisiana, USA (2018)

\bibitem{clef2018checkthat:task2}
Barr\'{o}n-Cede\~{n}o, A., Elsayed, T., Suwaileh, R., M\`{a}rquez, L.,
  Atanasova, P., Zaghouani, W., Kyuchukov, S., Da~San~Martino, G., Nakov, P.:
  Overview of the {CLEF-2018 CheckThat! Lab} on automatic identification and
  verification of political claims, {T}ask 2: Factuality. In: Cappellato, L.,
  Ferro, N., Nie, J.Y., Soulier, L. (eds.) CLEF 2018 Working Notes. Working
  Notes of CLEF 2018 - Conference and Labs of the Evaluation Forum. {CEUR}
  Workshop Proceedings, CEUR-WS.org, Avignon, France (2018)

\bibitem{castillo2011information}
Castillo, C., Mendoza, M., Poblete, B.: Information credibility on {T}witter.
  In: Proceedings of the 20th International Conference on World Wide Web. pp.
  675--684. WWW~'11, Hyderabad, India (2011)

\bibitem{cer2018universal}
Cer, D., Yang, Y., Kong, S.Y., Hua, N., Limtiaco, N., John, R.S., Constant, N.,
  Guajardo-Cespedes, M., Yuan, S., Tar, C., et~al.: Universal sentence encoder.
  arXiv preprint arXiv:1803.11175  (2018)

\bibitem{checkthat19:Coca}
Coca, L., Cusmuliuc, C.G., Iftene, A.: {CheckThat! 2019 UAICS}. In: CLEF 2019
  Working Notes. Working Notes of CLEF 2019 - Conference and Labs of the
  Evaluation Forum. {CEUR} Workshop Proceedings, CEUR-WS.org, Lugano,
  Switzerland (2019)

\bibitem{checkthat19:Dhar}
Dhar, R., Dutta, S., Das, D.: A hybrid model to rank sentences for
  check-worthiness. In: CLEF 2019 Working Notes. Working Notes of CLEF 2019 -
  Conference and Labs of the Evaluation Forum. {CEUR} Workshop Proceedings,
  CEUR-WS.org, Lugano, Switzerland (2019)

\bibitem{CheckThat:ECIR2019}
Elsayed, T., Nakov, P., Barr{\'o}n-Cede{\~{n}}o, A., Hasanain, M., Suwaileh,
  R., Da~San~Martino, G., Atanasova, P.: {CheckThat! at CLEF 2019}: Automatic
  identification and verification of claims. In: Azzopardi, L., Stein, B.,
  Fuhr, N., Mayr, P., Hauff, C., Hiemstra, D. (eds.) Advances in Information
  Retrieval. pp. 309--315. Springer International Publishing (2019)

\bibitem{checkthat19:Favano}
Favano, L., Carman, M., Lanzi, P.: {TheEarthIsFlat's} submission to {CLEF'19
  CheckThat!} challenge. In: CLEF 2019 Working Notes. Working Notes of CLEF
  2019 - Conference and Labs of the Evaluation Forum. {CEUR} Workshop
  Proceedings, CEUR-WS.org, Lugano, Switzerland (2019)

\bibitem{checkthat19:Gasior}
Gasior, J., Przyby{\l}a, P.: The {IPIPAN} team participation in the
  check-worthiness task of the {CLEF2019 CheckThat! Lab}. In: CLEF 2019 Working
  Notes. Working Notes of CLEF 2019 - Conference and Labs of the Evaluation
  Forum. {CEUR} Workshop Proceedings, CEUR-WS.org, Lugano, Switzerland (2019)

\bibitem{gencheva-EtAl:2017:RANLP}
Gencheva, P., Nakov, P., M\`{a}rquez, L., Barr\'{o}n-Cede\~{n}o, A., Koychev,
  I.: A context-aware approach for detecting worth-checking claims in political
  debates. In: Proceedings of the International Conference Recent Advances in
  Natural Language Processing. pp. 267--276. RANLP~'17, Varna, Bulgaria (2017)

\bibitem{checkthat19:Ghanem}
Ghanem, B., Glavaš, G., Giachanou, A., Ponzetto, S., Rosso, P., Rangel, F.:
  {UPV-UMA at CheckThat! Lab}: Verifying {A}rabic claims using cross lingual
  approach. In: CLEF 2019 Working Notes. Working Notes of CLEF 2019 -
  Conference and Labs of the Evaluation Forum. {CEUR} Workshop Proceedings,
  CEUR-WS.org, Lugano, Switzerland (2019)

\bibitem{hansen2019neural}
Hansen, C., Hansen, C., Alstrup, S., Grue~Simonsen, J., Lioma, C.: Neural
  check-worthiness ranking with weak supervision: Finding sentences for
  fact-checking. In: Companion Proceedings of the 2019 World Wide Web
  Conference. pp. 994--1000. WWW '19, San Francisco, USA (2019)

\bibitem{checkthat19:Hansen}
Hansen, C., Hansen, C., Simonsen, J., Lioma, C.: Neural weakly supervised fact
  check-worthiness detection with contrastive sampling-based ranking loss. In:
  CLEF 2019 Working Notes. Working Notes of CLEF 2019 - Conference and Labs of
  the Evaluation Forum. {CEUR} Workshop Proceedings, CEUR-WS.org, Lugano,
  Switzerland (2019)

\bibitem{checkthat19:Haouari}
Haouari, F., Ali, Z., Elsayed, T.: {bigIR at CLEF 2019}: Automatic verification
  of {A}rabic claims over the web. In: CLEF 2019 Working Notes. Working Notes
  of CLEF 2019 - Conference and Labs of the Evaluation Forum. {CEUR} Workshop
  Proceedings, CEUR-WS.org, Lugano, Switzerland (2019)

\bibitem{Hardalov2016}
Hardalov, M., Koychev, I., Nakov, P.: In search of credible news. In:
  Proceedings of the 17th International Conference on Artificial Intelligence:
  Methodology, Systems, and Applications. pp. 172--180. AIMSA~'16, Varna,
  Bulgaria (2016)

\bibitem{Hassan:15}
Hassan, N., Li, C., Tremayne, M.: Detecting check-worthy factual claims in
  presidential debates. In: Proceedings of the 24th ACM International
  Conference on Information and Knowledge Management. pp. 1835--1838. CIKM~'15,
  Melbourne, Australia (2015)

\bibitem{howard2018universal}
Howard, J., Ruder, S.: Universal language model fine-tuning for text
  classification. arXiv preprint arXiv:1801.06146  (2018)

\bibitem{NAACL2018:claimrank}
Jaradat, I., Gencheva, P., Barr\'on-Cede{\~n}o, A., M\`{a}rquez, L., Nakov, P.:
  {ClaimRank}: Detecting check-worthy claims in {A}rabic and {E}nglish. In:
  Proceedings of the 16th Annual Conference of the North American Chapter of
  the Association for Computational Linguistics. pp. 26--30. NAACL-HLT~'18, New
  Orleans, Louisiana, USA (2018)

\bibitem{jarvelin2002cumulated}
J{\"a}rvelin, K., Kek{\"a}l{\"a}inen, J.: Cumulated gain-based evaluation of
  {IR} techniques. ACM Transactions on Information Systems (TOIS)
  \textbf{20}(4),  422--446 (2002)

\bibitem{RANLP2017:clickbait}
Karadzhov, G., Gencheva, P., Nakov, P., Koychev, I.: We built a fake news \&
  click-bait filter: What happened next will blow your mind! In: Proceedings of
  the 2017 International Conference on Recent Advances in Natural Language
  Processing. pp. 334--343. RANLP~'17, Varna, Bulgaria (2017)

\bibitem{RANLP2017:factchecking:external}
Karadzhov, G., Nakov, P., M\`{a}rquez, L., Barr\'on-Cede{\~n}o, A., Koychev,
  I.: Fully automated fact checking using external sources. In: Proceedings of
  the International Conference on Recent Advances in Natural Language
  Processing. pp. 344--353. RANLP~'17, Varna, Bulgaria (2017)

\bibitem{ma2016detecting}
Ma, J., Gao, W., Mitra, P., Kwon, S., Jansen, B.J., Wong, K.F., Cha, M.:
  Detecting rumors from microblogs with recurrent neural networks. In:
  Proceedings of the 25th International Joint Conference on Artificial
  Intelligence. pp. 3818--3824. IJCAI~'16, New York, New York, USA (2016)

\bibitem{manning2010}
Manning, C.D., Raghavan, P., Sch\"{u}tze, H.: Introduction to Information
  Retrieval. Cambridge University Press, New York, NY, USA (2008)

\bibitem{mihaylova-etal-2019-semeval}
Mihaylova, T., Karadzhov, G., Atanasova, P., Baly, R., Mohtarami, M., Nakov,
  P.: {S}em{E}val-2019 task 8: Fact checking in community question answering
  forums. In: Proceedings of the 13th International Workshop on Semantic
  Evaluation. pp. 860--869. SemEval~'19, Minneapolis, Minnesota, USA (2019)

\bibitem{AAAI2018:factchecking}
Mihaylova, T., Nakov, P., M\`{a}rquez, L., Barr\'on-Cede{\~n}o, A., Mohtarami,
  M., Karadjov, G., Glass, J.: Fact checking in community forums. In:
  Proceedings of the 33rd AAAI Conference on Artificial Intelligence. pp.
  5309--5316. AAAI~'18, New Orleans, Louisiana, USA (2018)

\bibitem{mikolov-yih-zweig:2013:NAACL-HLT}
Mikolov, T., Yih, W.T., Zweig, G.: Linguistic regularities in continuous space
  word representations. In: Proceedings of the 2013 Conference of the North
  American Chapter of the Association for Computational Linguistics: Human
  Language Technologies. pp. 746--751. NAACL-HLT~'13, Atlanta, Georgia, USA
  (2013)

\bibitem{checkthat19:Mohtaj}
Mohtaj, S., Himmelsbach, T., Woloszyn, V., Möller, S.: The {TU-Berlin} team
  participation in the check-worthiness task of the {CLEF-2019 CheckThat! Lab}.
  In: CLEF 2019 Working Notes. Working Notes of CLEF 2019 - Conference and Labs
  of the Evaluation Forum. {CEUR} Workshop Proceedings, CEUR-WS.org, Lugano,
  Switzerland (2019)

\bibitem{mukherjee2015leveraging}
Mukherjee, S., Weikum, G.: Leveraging joint interactions for credibility
  analysis in news communities. In: Proceedings of the 24th ACM International
  on Conference on Information and Knowledge Management. pp. 353--362. CIKM
  '15, Melbourne, Australia (2015)

\bibitem{clef2018checkthat:overall}
Nakov, P., Barr\'{o}n-Cede\~{n}o, A., Elsayed, T., Suwaileh, R., M\`{a}rquez,
  L., Zaghouani, W., Atanasova, P., Kyuchukov, S., Da~San~Martino, G.: Overview
  of the {CLEF-2018 CheckThat! Lab} on automatic identification and
  verification of political claims. In: Bellot, P., Trabelsi, C., Mothe, J.,
  Murtagh, F., Yun~Nie, J., Soulier, L., Sanjuan, E., Cappellato, L., Ferro, N.
  (eds.) Proceedings of the Ninth International Conference of the CLEF
  Association: Experimental IR Meets Multilinguality, Multimodality, and
  Interaction. Lecture Notes in Computer Science, Springer, Avignon, France
  (September 2018)

\bibitem{nguyen2018interpretable}
Nguyen, A.T., Kharosekar, A., Lease, M., Wallace, B.: An interpretable joint
  graphical model for fact-checking from crowds. In: Proceedings of the 32nd
  AAAI Conference on Artificial Intelligence. pp. 1511--1518. AAAI~'18, New
  Orleans, LA, USA (2018)

\bibitem{Nie19}
Nie, Y., Chen, H., Bansal, M.: Combining fact extraction and verification with
  neural semantic matching networks. In: Proceedings of the 33rd AAAI
  Conference on Artificial Intelligence. AAAI~'19, Honolulu, Hawaii, USA (2019)

\bibitem{popat2016credibility}
Popat, K., Mukherjee, S., Str\"{o}tgen, J., Weikum, G.: Credibility assessment
  of textual claims on the web. In: Proceedings of the 25th ACM International
  Conference on Information and Knowledge Management. pp. 2173--2178. CIKM '16,
  Indianapolis, Indiana, USA (2016)

\bibitem{popat2018declare}
Popat, K., Mukherjee, S., Yates, A., Weikum, G.: {D}e{C}lar{E}: Debunking fake
  news and false claims using evidence-aware deep learning. In: Proceedings of
  the 2018 Conference on Empirical Methods in Natural Language Processing. pp.
  22--32. EMNLP~'18, Brussels, Belgium (2018)

\bibitem{rubin2015deception}
Rubin, V.L., Chen, Y., Conroy, N.J.: Deception detection for news: three types
  of fakes. In: Proceedings of the 78th ASIS\&T Annual Meeting: Information
  Science with Impact: Research in and for the Community. p.~83. American
  Society for Information Science (2015)

\bibitem{shu2017fake}
Shu, K., Sliva, A., Wang, S., Tang, J., Liu, H.: Fake news detection on social
  media: A data mining perspective. ACM SIGKDD Explorations Newsletter
  \textbf{19}(1),  22--36 (2017)

\bibitem{checkthat19:Su}
Su, T., Macdonald, C., Ounis, I.: Entity detection for check-worthiness
  prediction: {Glasgow Terrier at CLEF CheckThat!} 2019. In: CLEF 2019 Working
  Notes. Working Notes of CLEF 2019 - Conference and Labs of the Evaluation
  Forum. {CEUR} Workshop Proceedings, CEUR-WS.org, Lugano, Switzerland (2019)

\bibitem{thorne2018fever}
Thorne, J., Vlachos, A., Christodoulopoulos, C., Mittal, A.: {FEVER}: a
  large-scale dataset for {F}act {E}xtraction and {VER}ification. In:
  Proceedings of the 2018 Conference of the North American Chapter of the
  Association for Computational Linguistics: Human Language Technologies. pp.
  809--819. NAACL-HLT~'18, New Orleans, LA, USA (2018)

\bibitem{checkthat19:Touahri}
Touahri, I., Mazroui, A.: Automatic identification and verification of
  political claims. In: CLEF 2019 Working Notes. Working Notes of CLEF 2019 -
  Conference and Labs of the Evaluation Forum. {CEUR} Workshop Proceedings,
  CEUR-WS.org, Lugano, Switzerland (2019)

\bibitem{Yasser2018:CIKM}
Yasser, K., Kutlu, M., Elsayed, T.: Re-ranking web search results for better
  fact-checking: A preliminary study. In: Proceedings of 27th ACM International
  Conference on Information and Knowledge Management. pp. 1783--1786. CIKM~'19,
  Turin, Italy (2018)

\bibitem{yoneda2018ucl}
Yoneda, T., Mitchell, J., Welbl, J., Stenetorp, P., Riedel, S.: {UCL} machine
  reading group: Four factor framework for fact finding ({HexaF}). In:
  Proceedings of the First Workshop on Fact Extraction and VERification. pp.
  97--102. FEVER~'18, Brussels, Belgium (2018)

\bibitem{zhu2016computing}
Zhu, G., Iglesias, C.A.: Computing semantic similarity of concepts in knowledge
  graphs. IEEE Transactions on Knowledge and Data Engineering  \textbf{29}(1),
  72--85 (2016)

\bibitem{zhu2015sematch}
Zhu, G., Iglesias~Fernandez, C.A.: Sematch: semantic entity search from
  knowledge graph. In: Joint Proceedings of the 1st International Workshop on
  Summarizing and Presenting Entities and Ontologies and the 3rd International
  Workshop on Human Semantic Web Interfaces. SumPre-HSWI@ESWC~'2015, Portorož,
  Slovenia (2015)

\bibitem{zubiaga2016analysing}
Zubiaga, A., Liakata, M., Procter, R., Hoi, G.W.S., Tolmie, P.: Analysing how
  people orient to and spread rumours in social media by looking at
  conversational threads. PloS one  \textbf{11}(3),  e0150989 (2016)

\end{thebibliography}
\end{document}